\documentclass[runningheads]{llncs}
\usepackage{graphicx}

\usepackage{tikz}
\usepackage{comment}
\usepackage{amsmath,amssymb} 
\usepackage{color}
\usepackage{breakcites}

\usepackage[accsupp]{axessibility}  

\usepackage{bm}
\usepackage{url}
\usepackage{pifont}
\usepackage{bbding}
\usepackage{comment}
\usepackage{caption}
\usepackage{amsmath}
\usepackage{amssymb}
\usepackage{hyperref}
\usepackage{pdfpages}
\usepackage{multirow}
\usepackage{makecell}
\usepackage{amsfonts}
\usepackage{booktabs}
\usepackage{subfigure}
\usepackage[misc]{ifsym}
\usepackage{pifont}

\def\ie{{\em i.e. }}
\def\eg{{\em e.g. }}

\begin{document}
\pagestyle{headings}
\mainmatter
\def\ECCVSubNumber{4001}  

\title{PromptDet: Towards Open-vocabulary Detection using Uncurated Images} 


\titlerunning{PromptDet: Towards Open-vocabulary Detection}
%
\author{Chengjian Feng\inst{1} \and
Yujie Zhong\inst{1} \and
Zequn Jie\inst{1} \and
Xiangxiang Chu\inst{1} \\
Haibing Ren\inst{1} \and
Xiaolin Wei\inst{1} \and
Weidi Xie\inst{2, \text{ \Letter}} \and
Lin Ma\inst{1}
}
\authorrunning{C. Feng et al.}
%
\institute{$^1$Meituan Inc. \hspace{3pt} $^2$Shanghai Jiao Tong University
}
\maketitle

\newcommand{\xmark}{\ding{55}}%

\begin{abstract}
The goal of this work is to establish a scalable pipeline for 
expanding an object detector towards novel/unseen categories, 
using {\em zero manual annotations}.
To achieve that, we make the following four contributions:
(i) in pursuit of generalisation, 
we propose a two-stage open-vocabulary object detector, 
where the class-agnostic object proposals are classified with a text encoder from pre-trained visual-language model;
(ii) To pair the visual latent space~(of RPN box proposals) with that of the pre-trained text encoder, we propose the idea of {\em regional prompt learning} to align the textual embedding space with regional visual object features;
(iii) To scale up the learning procedure towards detecting a wider spectrum of objects, we exploit the available online resource via a novel self-training framework, which allows to train the proposed detector on a large corpus of noisy uncurated web images. 
Lastly,  
(iv) to evaluate our proposed detector, termed as {\bf PromptDet},
we conduct extensive experiments on the challenging LVIS and MS-COCO dataset. PromptDet shows superior performance over existing approaches with {\em fewer additional training images} and {\em zero manual annotations} whatsoever.
Project page with code: \url{https://fcjian.github.io/promptdet}.
\end{abstract}

\section{Introduction}
Object detection has been one of the most widely researched problems in computer vision, 
with the goal of simultaneously localising and categorising objects in the image. 
In the recent literature, 
the detection community has witnessed tremendous success by training on large-scale datasets, {\em e.g.}~PASCAL VOC~\cite{everingham2015pascal}, MS-COCO~\cite{lin2014microsoft}, 
with objects of certain category being exhaustively annotated with bounding box and category labels.
However, the scalability of such training regime is clearly limited,
as the model can only perform well on a closed
and small set of categories for which large-scale data is easy to collect and annotate.

On the other hand, 
the recent large-scale visual-language pre-training has shown tremendous success in open-vocabulary image classification, 
which opens up the opportunity for expanding the vocabulary a detector can operate on. 
In specific, these visual-language models (for example, CLIP~\cite{radford2021learning} and ALIGN~\cite{jia2021scaling}) are often trained on billion-scale noisy image-text pairs, 
with noise contrastive learning,
and has demonstrated a basic understanding on `what' generally are the salient objects in an image.
However, training detectors in the same manner, 
{\em i.e.}~using image-text pairs, clearly poses significant challenge on scalability, as it would require the captions to not only include semantics~(\ie `what'), but also the spatial information~(\ie `where') of the objects.
As a result, the community considers a slightly conservative scenario in open-vocabulary object detection~\cite{gu2021open,zhou2022detecting}:  
{\bf given an existing object detector trained on abundant data for some base categories, we wish to expand the detector's ability to localise and recognise novel categories, with minimal human effort}.

\begin{figure*}[!tb]
\centering
\includegraphics[width=\textwidth]{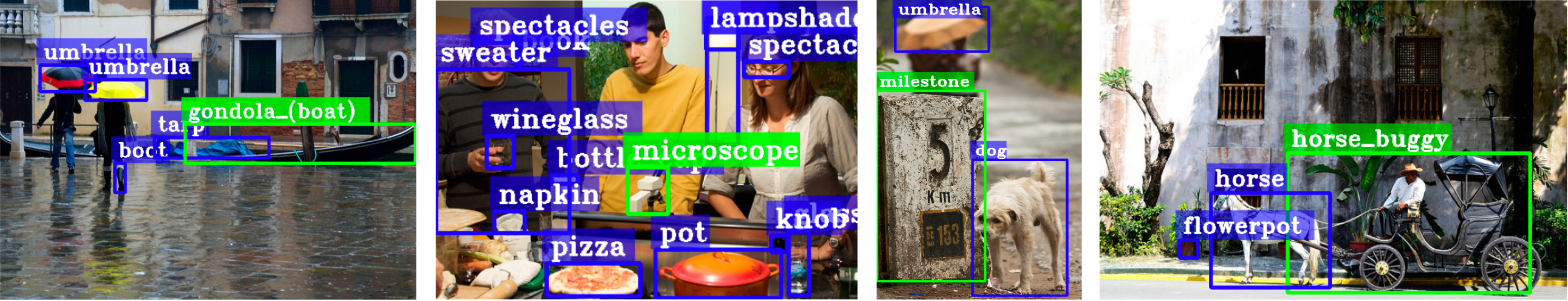}
\caption{
The proposed PromptDet is  
a framework for expanding the vocabulary of an object detector without human annotation. The figure depicts detection examples produced by our model on LVIS validation set, with blue and green boxes denoting the objects from \textcolor{blue}{\textbf{base}} and \textcolor{green}{\textbf{novel}} categories respectively. 
Despite no ground truth annotation is provided for the novel categories, 
PromptDet is still able to reliably localise and recognise these objects with high accuracy. 
\vspace{-.2cm}
}
\label{qualitative_row}
\end{figure*}

This paper describes a simple idea for pairing the visual latent space with a pre-trained language encoder,
\eg inheriting the CLIP's text encoder as a `classifier' generator, 
and only train the detector's visual backbone and class-agnostic region proposal. 
The novelty of our approach is in the two steps for aligning the visual and textual latent spaces.
Firstly, we propose to learn a certain number prompt vectors on the textual encoder side, termed as regional prompt learning~(RPL), 
such that its latent space can be transformed to better pair with the visual embeddings from the object-centric feature.
Secondly, by leveraging a large corpus of uncurated web images, 
we further iteratively optimise the prompt vectors by retrieving a set of candidate images from the Internet, and self-train the detector with the pseudo-labels generated on the sourced candidate images.
The resultant detector is named {\bf PromptDet}.
Experimentally, despite the noise in the image candidates,
such self-training regime has shown a noticeable improvement on the open-vocabulary generalisation, 
especially on the categories where no box annotations are available.
Detection examples produced by PromptDet are illustrated in Figure~\ref{qualitative_row}.

To summarise, we make the following contributions: 
(i) We investigate the problem of open-vocabulary detection based on a simple idea, namely, equipping a standard two-stage object detector with a frozen textual encoder from a pre-trained visual-language model. 
(ii) We propose a regional prompt learning approach for transforming the embedding space of the text encoder to better fit the object-centric feature proposed by the RPN of the detector.
(iii) We introduce a novel learning framework, which allows to iteratively update the prompts and source high-quality external images from the web using the updated prompts, and finally self-train the detector. 
(iv) PromptDet substantially outperforms previous state-of-the-art on the LVIS~\cite{gupta2019lvis} and MS-COCO~\cite{lin2014microsoft}, despite only using uncurated web images and much fewer training costs, {\em i.e.}~smaller image resolution and fewer epochs.

\section{Related Work}
\par{\noindent \bf Object Detection.}
Generally speaking, 
modern object detection frameworks can be divided into two-stage~\cite{he2017mask,ren2015faster,feng2021exploring} and one-stage ones~\cite{lin2017focal,tian2019fcos,feng2021tood,zhong2020representation}. 
The two-stage detectors first generate a set of region proposals, 
and then classify and refine these proposals. 
In contrast, the one-stage detectors directly predict the category and bounding box at each location. 
The majority of existing detectors require a large number of training data with bounding box annotations, and can only recognise a fixed set of categories that are present in the training data. 
Recently, several works develop few-shot detection~\cite{kang2019few,fan2020few, Kaul22} and zero-shot detection~\cite{gu2021open,zhou2022detecting} to relax the restriction from expensive data annotations. 
In specific, few-shot detection aims to detect novel categories by adjusting the detector with one or few annotated samples. 
Zero-shot detection aims to identify novel categories without any additional examples.\\[-6pt]

\par{\noindent \bf Open-vocabulary Object Detection.}
In the recent literature, open-vocabulary detection has attracted increasingly more interest within the community.
The goal is to detect objects beyond a closed set,
in~\cite{bansal2018zero}, the authors propose to replace the last classification layer with language embeddings of the class names. 
\cite{Li19,Rahman20} introduce the external text information while computing the classifier embedding. 
OVR-CNN~\cite{Zareian21} trains the detector on image-text pairs with contrastive learning. 
ViLD~\cite{gu2021open} and ZSD-YOLO~\cite{xie2021zsd} propose to explicitly distill the knowledge from the pre-trained CLIP visual embedding into the visual backbone of a Faster RCNN.
One closely related work is the Detic~\cite{zhou2022detecting},
which seeks to self-train the detector on ImageNet21K, 
to expand the vocabulary of detector.
Nonetheless, it still requires a tremendous amount of human effort for annotating these ImageNet21K images.
In contrast, 
our proposed self-training framework poses less limitations, 
and enables to directly train on uncurated web images.\\[-6pt]

\par{\noindent \bf Zero-shot Learning.}
Zero-shot learning aims to transfer the learned knowledge from some seen object classes to novel classes. In object recognition, early works exploit the visual attribution such as class hierarchy, 
class similarity and object parts to generalize from seen classes to unseen classes~\cite{rohrbach2011evaluating,akata2016multi,zhao2017open,elhoseiny2017link,ji2018stacked}. 
Other line of research learns to map visual samples and the semantic descriptors to a joint embedding space, 
and compute the similarity between images and free form texts in the embedding space~\cite{frome2013devise,cacheux2019modeling}. \\[-6pt]

\par{\noindent \bf Vision-Language Pre-training.}
In computer vision, 
joint visual-textual learning has been researched for a long time. In the early work from Mori et. al.~\cite{Mori99}, 
connections between image and words in paired text documents were first explored, 
\cite{Weston11} learnt a joint image-text embedding for the case of class name annotations. 
In the recent literature, CLIP~\cite{radford2021learning} and ALIGN~\cite{jia2021scaling} collect a million/billion-scale image-caption pairs from the Internet, and jointly train an image encoder and a text encoder with simple noise contrastive learning, 
which has shown to be extremely effective for a set of downstream tasks, such as zero-shot image classification.

\section{Methodology}
In this paper, 
we aim to expand the vocabulary of a standard two-stage object detector, 
to localise and recognise objects from novel categories with minimal manual effort.
This section is organized as follows, 
we start by introducing the basic blocks for building an open-vocabulary detector, 
that can detect objects of arbitrary category, beyond a closed set;
In Section~\ref{sec:naive}, 
we describe the basic idea for pairing the visual backbone with a frozen language model, 
by inheriting CLIP's text encoder as a classifier generator; 
In Section~\ref{sec:prompt}, 
to encourage alignment between the object-centric visual representation and textual representation, we introduce the regional prompt learning (RPL);
In Section~\ref{visual-representation}, 
we introduce an iterative learning scheme,
that can effectively leverage the uncurated web images, 
and source high-quality candidate images of novel categories.
As a consequence, our proposed open-vocabulary detector, termed as {\bf PromptDet}, 
can be self-trained on these candidate images in a scalable manner.

\subsection{Open Vocabulary Object Detector}
\label{sec:baseline}
Here, we consider the same problem setup as in~\cite{gu2021open}. 
Assuming we are given an image detection dataset, $\mathcal{D}_{\text{train}}$, 
with exhaustive annotations on a set of base categories, $\mathcal{C}_{\text{train}} = \mathcal{C}_{\text{base}}$,
{\em i.e.}~$\mathcal{D}_{\text{train}} = \{(I_1, y_1), \dots, (I_n, y_n)\}$, 
where $I_i \in \mathbb{R}^{H \times W \times 3}$ refers to the $i$-th image, and $y_i = \{(b_i, c_i)\}^m$ denotes the coordinates~($b^k_i \in \mathbb{R}^4$) 
and category label~($c^k_i \in \mathbb{R}^{\mathcal{C}_{\text{base}}}$) for a total of $m$ objects in such image.
The goal is to train an object detector that can successfully operate on a test set, $\mathcal{D}_{test}$,
with objects beyond a closed set of base categories,
{\em i.e.}~$\mathcal{C}_{\text{test}} = \mathcal{C}_{\text{base}} \cup \mathcal{C}_{\text{novel}}$, 
thus termed as an open-vocabulary detector.
In particular, we conduct the experiments on LIVS dataset~\cite{gupta2019lvis},
and treat the union of common and frequent classes as {\em base} categories, and the rare classes as {\em novel} categories.

Generally speaking, a popular two-stage object detector, 
for example, Mask-RCNN,
is consisted of a visual backbone encoder, 
a region proposal network~(RPN) and classification module:
\begin{align}
    \{\hat{y}_1, \dots, \hat{y}_n\} = \mathrm{\Phi}_{\text{CLS}} \circ
        \mathrm{\Phi}_{\text{RPN}} \circ \mathrm{\Phi}_{\text{ENC}}(I)
\end{align}
Constructing an open-vocabulary detector would therefore require to solve two subsequent problems: 
(1) to effectively generate class-agnostic region proposals,
and (2) to accurately classify each of these proposed regions beyond a close set of visual categories, {\em i.e.}~open-vocabulary classification.\\[-8pt]

\begin{figure*}[t]
		\centering
		\includegraphics[width=\textwidth]{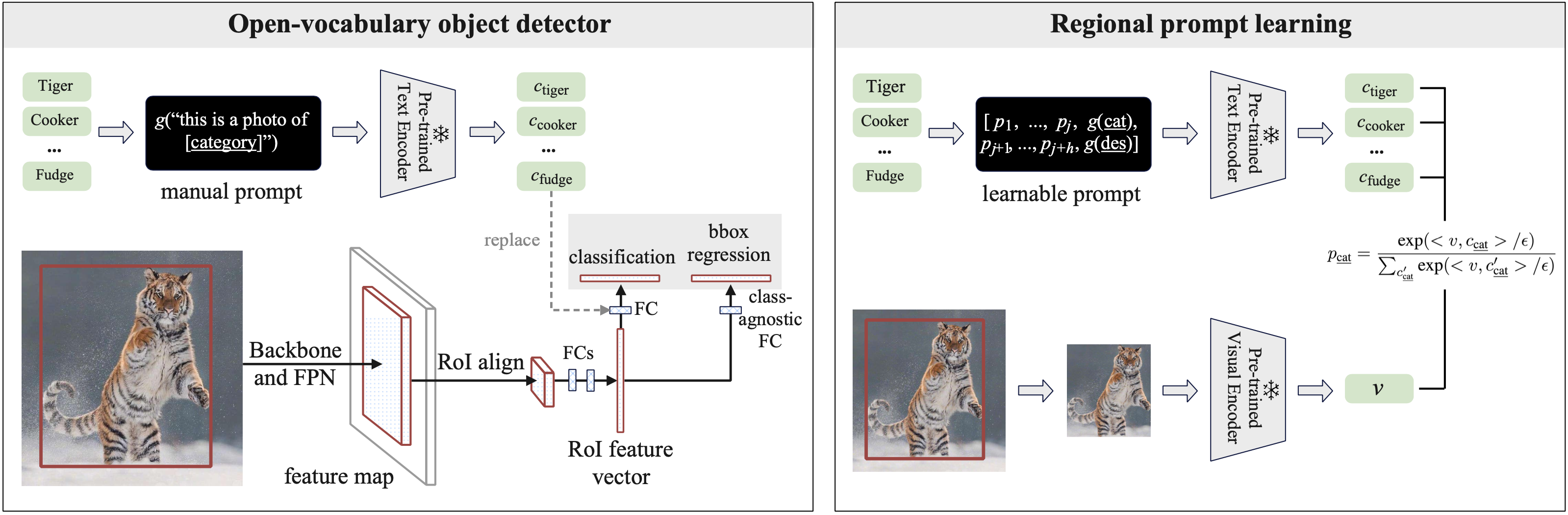}
		\caption{
		\textbf{Left}: the proposed open-vocabulary object detector.
		We inherit the category embeddings as an open-vocabulary classifier, and train the visual backbone to align with the classifier. 
		\textbf{Right}: the procedure for the off-line regional prompt learning.
		We take crops for all base categories, and use their visual embeddings to optimise the learnable prompt
		(Section~\ref{sec:prompt} has detailed description).
		}
		\label{overview}
		\vspace{-0.2cm}
\end{figure*}

{\noindent \bf Class-agnostic region proposal networks~($\mathrm{\Phi}_{\text{RPN}}$): }
refers to the ability of proposing all regions that are likely to have objects, regardless of their categories.
Here, we parametrise the anchor classification,
bounding box regression and mask prediction in a class-agnostic manner, 
{\em i.e.}~sharing parameters for all the categories.
This is also in line with the discovery in recent work~\cite{Kaul22, zhou2022detecting}. \\[-8pt]

{\noindent \bf Open-vocabulary classification~($\mathrm{\Phi}_{\text{CLS}}$): } 
aims to categorise the visual object beyond a fixed-size vocabulary.
We make the assumption that, 
there exists a common latent space between vision and natural language, 
classifying any visual object can thus be achieved by looking for its closest embedding in the language latent space,
for example, to classify a region as ``\textit{almond}'' or ``\textit{dog}'',
the classification probability for being ``\textit{almond}'' can be computed:
\begin{align}
&c_{\text{almond}} = \phi_{\text{text}}(g(\text{``this is a photo of [\underline{almond}]''})) \\
&c_{\text{dog}} = \phi_{\text{text}}(g(\text{``this is a photo of [\underline{dog}]''})) \\
&p_{\text{almond}} = \frac{\text{exp}(<v, c_{\text{almond}}>/\epsilon)}
                    {\text{exp}(<v, c_{\text{almond}}>/\epsilon) + \text{exp}(<v, c_{\text{dog}}>/\epsilon)} 
\end{align}
where $v \in \mathbb{R}^{D}$ denotes the ROI pooled features from region proposal network, 
$g(\cdot)$ refers to a simple tokenisation procedure, 
with no trainable parameters,
$\phi_{\text{text}}$ denotes a hyper-network that maps the natural language to its corresponding latent embedding,
note that, the input text usually requires to use a template with manual prompts, 
{\em e.g.}~``this is a photo of [\underline{category}]'',
which converts the classification tasks into the same format as that used during pre-training. As both visual and textual embedding have been L2 normalised, 
a temperature parameter $\epsilon$ is also introduced.
The visual backbone is trained by optimising the classification loss, to pair the regional visual embedding and its textual embedding of the corresponding category.\\[-8pt]

{\noindent \bf Discussion: }
Despite the simplicity in formulating an open-vocabulary detector,
training such models would suffer from great challenges, 
due to the lack of exhaustive annotations for large-scale dataset.
Until recently,
the large-scale visual-language models, 
such as CLIP and ALIGN,
have been trained to align the latent space between vision and language, using simple noise contrastive learning at the image level.
Taking benefit from the rich information in text descriptions, 
{\em e.g.}~actions, objects, human-object interactions, 
and object-object relationships,
these visual-language models have demonstrated remarkable `zero-shot' generalisation for various image classification tasks,
which opens up the opportunity for expanding the vocabulary of an object detector.

\subsection{Na\"ive Alignment via Detector Training}
\label{sec:naive}
In this section,
we aim to train an open-vocabulary object detector (based on Mask-RCNN) on $\mathcal{D}_{\text{train}}$, 
{\em i.e.}~only {\em base} categories, 
by optimising the visual backbone and the class-agnostic RPN to align with the object category classifier, 
that is inherited from the pre-trained frozen text encoder from CLIP, as shown in Figure~\ref{overview}~(left).
Note that, such training regime has also been investigated in several previous work, for example, \cite{bansal2018zero,gu2021open}.

However, as indicated by our experiments, 
na\"ively aligning the visual latent space to textual ones only yields very limited open-vocabulary detection performance.
We conjecture that the poor generalisation mainly comes from three aspects:
Firstly, computing the category embedding with only class name is sub-optimal, as they may not be precise enough to describe a visual concept, leading to the lexical ambiguity, For example, 
``\textit{almond}'' either refers to an edible oval nut with a hard shell or the tree that it grows on; Secondly, the web images for training CLIP are scene-centric, 
with objects occupying only a small portion of the image, 
whereas the object proposals from RPNs often closely localise the object,
leading to an obvious domain gap on the visual representation; 
Thirdly, the base categories used for detector training are significantly less diverse than those used for training CLIP, 
thus, may not be sufficient to guarantee a generalisation towards novel categories. In the following sections, we propose a few simple steps to alleviate the above issues.

\subsection{Alignment via Regional Prompt Learning}
\label{sec:prompt}
Comparing with the scene-centric images used for training CLIP, 
the output features from RPNs are local and object-centric.
Na\"ively aligning the regional visual representation to the {\em frozen} CLIP text encoder would therefore encourage each proposal to capture more context than it is required. 
To this end, we propose a simple idea of {\bf regional prompt learning} (RPL), 
steering the textual latent space to better fit object-centric images.

Specifically, while computing the category classifier or embedding, 
we prepend and append a sequence of learnable vectors to the textual input,
termed as `continuous prompt vectors'.
These prompt vectors do not correspond to any real concrete words, 
and will be attended at the subsequent layers as if they were a sequence of `virtual tokens'.
Additionally, 
we also include more detailed description into the prompt template to alleviate the lexical ambiguity, for instance, \{\underline{category}: {\em ``almond''}, 
\underline{description}: {\em ``oval-shaped edible seed of the almond tree''}\}.
Note that, 
the description can often be easily sourced from Wikipedia or meta data from the dataset. 
The embedding for each individual category can thus be generated as:
\begin{align}
	c_{\text{almond}} = \phi_{\text{text}}([p_1, \dots, p_j, g(\underline{\text{category}}), p_{j+1} \dots, p_{j+h}, g(\underline{\text{description}})])
\end{align}
where $p_{i}$ ($i \in \{1, 2, ..., j + h\}$) denote the learnable prompt vectors with the same dimension as word embedding, 
$[\underline{\text{category}}]$ and $[\underline{\text{description}}]$ are calculated by tokenising the category name and detailed description. 
As the learnable vectors are class-agnostic, 
and shared for all categories, 
they are expected to be transferable to novel categories after training.\\[-8pt]

{\noindent \bf Optimising prompt vectors.}
To save computations, 
we consider to learn the prompt vectors in an off-line manner,
specifically, 
we take the object crops of {\em base} categories from LVIS,
resize them accordingly and pass through the frozen CLIP visual encoder, 
to generate the image embeddings.
To optimise the prompt vectors, 
we keep both the visual and textual encoder frozen, 
and only leave the learnable prompt vectors to be updated, 
with a standard cross-entropy loss to classify these image crops. 
The process of RPL is displayed in Figure~\ref{overview}~(right).\\[-8pt]

{\noindent \bf Discussion.}
With the proposed RPL, 
the textual latent space is therefore re-calibrated 
to match the object-centric visual embeddings.
Once trained, we can re-compute all the category embeddings,
and train the visual backbone to align with the prompted text encoder,
as described in Figure~\ref{overview}~(left).
In our experiments, we have confirmed the effectiveness of RPL in Section~\ref{sec:ablation}, which indeed leads noticeable improvements on the open-vocabulary generalisation.

\subsection{PromptDet: Alignment via Self-training}
\label{visual-representation}
\begin{figure*}[t]
	\centering
	\includegraphics[width=.98\textwidth]{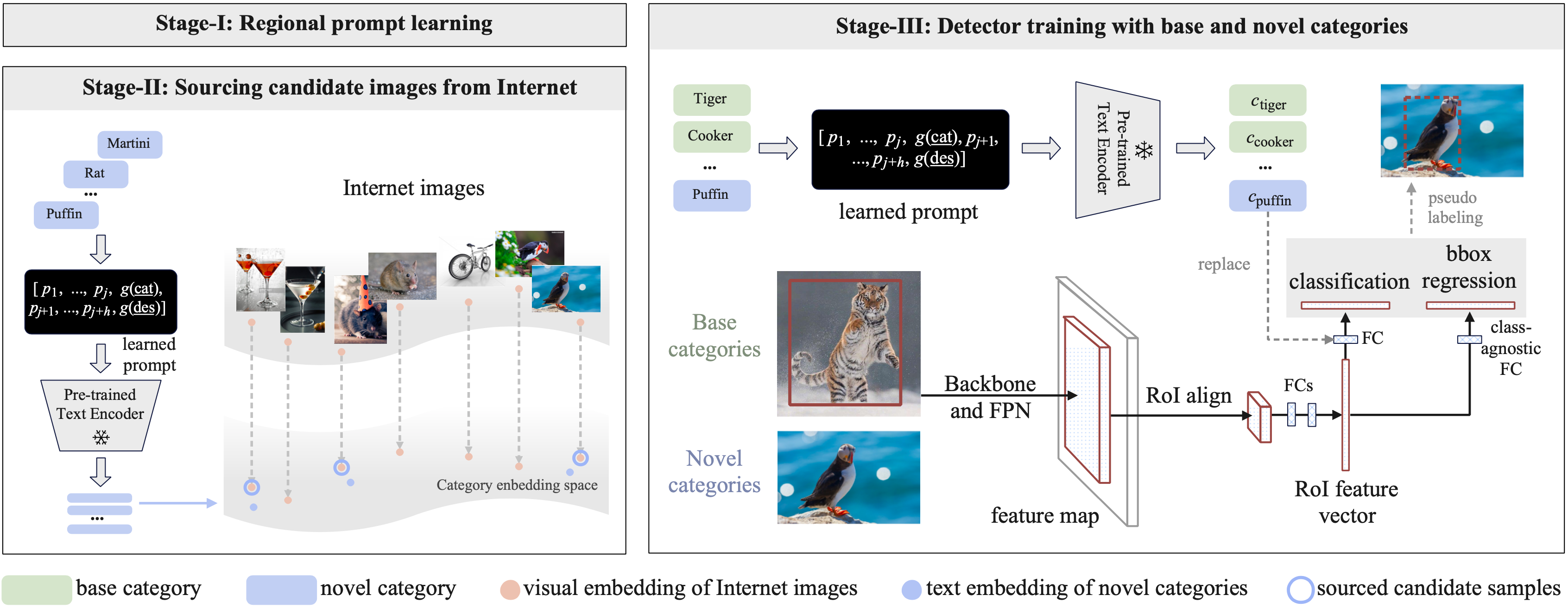}
	\caption{ \footnotesize Illustration of the self-training framework. 
	\textbf{Stage-I}: we use the base categories to learn regional prompts, 
	as already demonstrated in Figure~\ref{overview} (right). 
	\textbf{Stage-II}: we source and download the Internet images with the learned prompt. \textbf{Stage-III}: we self-train the detector with both LVIS images of base categories and the sourced images of novel categories. Note that, the prompt learning and image sourcing can be iteratively conducted to better retrieve relevant images.
	}
	\vspace{-0.3cm}
	\label{iterative}
\end{figure*}

Till here, we have obtained an open-vocabulary object detector by 
aligning the visual backbone to prompted text encoder. 
However, RPL has only exploited limited visual diversity,
{\em i.e.}~only with base categories.
In this section, we unleash such limitation
and propose to leverage the large-scale, uncurated, 
noisy web images to further improve the alignment.
Specifically, as shown in Figure~\ref{iterative}, 
we describe a learning framework that iterates the procedure of RPL and candidate images sourcing, followed by generating pseudo ground truth boxes,
and self-training the open-vocabulary detector.\\[-8pt]

{\noindent \bf Sourcing candidate images. }
We take the LAION-400M dataset as an initial corpus of images,
with the visual embeddings pre-computed by CLIP's visual encoder.
To acquire candidate images for each category, 
we compute the similarity score between the visual embedding and the category embedding, which are computed with the learnt regional prompt.
We keep the images with highest similarity~(an ablation study on the selection of the number of the images has been conducted in Section~\ref{sec:ablation}).
As a consequence, 
an additional set of images is constructed with both {\em base} and {\em novel} categories, with no ground truth bounding boxes available, 
{\em e.g.}~$\mathcal{D}_{\text{ext}} =  \{(I_\text{ext})_{i}\}_{i=1}^{|\mathcal{D}_{\text{ext}}|}$.\\[-8pt]

{\noindent \bf Iterative prompt learning and image sourcing. }
Here, we can alternate the procedure of the regional prompt learning (Figure~\ref{iterative} Stage-I) and sourcing Internet images with the learned prompt with high precision (Figure~\ref{iterative} Stage-II).
Experimentally, 
such iterative sourcing procedure has shown to be beneficial for mining object-centric images with high precision.
It enables to generate more accurate pseudo ground truth boxes
and, as a result, largely improves the detection performance on novel categories after self-training.\\[-8pt]

{\noindent \bf Bounding box generation. }
For each image in $\mathcal{D}_{\text{ext}}$,  
we run the inference with our open-vocabulary detector.
Since these sourced candidate images are often object-centric, 
the output object proposals from class-agnostic RPN usually guarantee a decent precision and recall. 
We retain the top-K proposals with max objectness scores~(experiments are conducted on selecting the value of K), 
then keep the box with the maximal classification score as the pseudo ground truth for each image.
Note that, despite an image may contain multiple objects of interest,
we only pick the one box as pseudo ground truth. 
Overall, such procedure would successfully mine a large set of previously unlabeled instances with pseudo ground truth,
which are later used for re-training the visual backbone and RPN~(including regression head) in Mask-RCNN, 
effectively resembling the self-training procedure.\\[-8pt]

{\noindent \bf Discussion. }
Recent Detic~\cite{zhou2022detecting} also attempts to train an open-vocabulary detector by exploiting external data, 
our proposed approach differs in three major aspects:
(1) in Detic, the ImageNet21K is used as the initial image corpus, 
which have already been well-curated with manual annotations, 
in contrast, we advocate more challenging and scalable scenario, 
with all external images uncurated;
(2) Detic uses a heuristic to pseudo-label the bounding boxes,
\ie to always pick the max-sized proposal;
while in our case, we choose the box with most confident prediction;
(3) our image sourcing and self-training can be iteratively conducted, 
and shown to lead significant performance boost, as indicated in Section~\ref{sec:ablation}.

\section{Experiment}

\subsection{Dataset \& Evaluation Metrics}
Here, we describe the open-vocabulary detection setup on LVIS~\cite{gupta2019lvis},
more details for MS-COCO benchmark can be found in supplementary material.\\[-8pt]

{\noindent \bf LVIS. }
The latest LVIS v1.0~\cite{gupta2019lvis} contains 1203 categories with both bounding box and instance mask annotations.
The categories are divided into three groups based on the number of the images that each category appears in the $train$ set: 
rare~(1-10 images), common~(11-100 images), and frequent~($>$100 images).
We follow the same problem setting as in ViLD~\cite{gu2021open} and Detic~\cite{zhou2022detecting}, 
where the frequent and common classes are treated as base categories~($\mathcal{C}_{base}$),
and the rare classes as the novel categories~($\mathcal{C}_{novel}$). \\[-8pt]


{\noindent \bf LAION-400M and LAION-Novel. }
For self-training, we also use an external dataset, LAION-400M~\cite{schuhmann2021laion},
which consists of 400 million image-text pairs filtered by pre-trained CLIP. 
It provides the pre-computed CLIP embedding for all the images, 
and we search for the images by using its 64G \emph{knn} indices and download about 300 images for each novel category, as illustrated by Stage-II in Figure~\ref{iterative}. 
We refer to this subset of LAION-400M as LAION-novel.

While training an initial open-vocabulary object detector, we use LVIS-base.
For self-training, we use a combination of LVIS-base and LAION-novel datasets, we summarise the dataset statistics in Table~\ref{dataset}.
For evaluation on LVIS v1.0 $minival$ set,
we mainly consider the mask Average Precision for novel categories, 
{\em i.e.}~AP$_{novel}$. However, to complete the AP metric, 
we also report AP$_{c}$~(for common classes) and AP$_{f}$~(for frequent classes). 
Lastly, the mask Average Precision for all categories is denoted by AP, 
which is computed as the mean of all the APs ranging from 0.5 to 0.95 IoU threshold (in a step of 0.05).

\vspace{-5pt}
\begin{table}[!htb]
\begin{center}
\vspace{-0.5cm}
\caption{A summary of dataset statistics. 
The numbers in bracket refer to the number of base and novel categories.}
\small
\setlength\tabcolsep{1.4pt}
\begin{tabular}{lccccc}
\hline\noalign{\smallskip}
Dataset & Train & Eval. & Definition & \#Images & \#Categories \\ \toprule
LVIS & -- & -- & original LVIS dataset & 0.1M & 1203 \\ 
LAION-400M & -- & -- & image-text pairs filtered by CLIP & 400M & unlabeled \\ \midrule
LVIS-base & $\checkmark$  & \xmark & common and frequent categories & 0.1M & 866 \\
LAION-novel & $\checkmark$ & \xmark & image subset of novel categories & 0.1M & 337 (noisy) \\ \midrule
LVIS $minival$ & \xmark & $\checkmark$ & standard LVIS validation set  & 20K & 1203~(866+337) \\
\bottomrule
\end{tabular}
\vspace{-1cm}
\label{dataset}
\end{center}
\end{table}

\subsection{Implementation details}
{\noindent \bf Detector training.}
We conduct all the experiments using Mask-RCNN~\cite{he2017mask} with a ResNet-50-FPN backbone.
Similar to Detic~\cite{zhou2022detecting}, we use $sigmoid$ activation and binary cross-entropy loss for classification.
We adopt the Stochastic Gradient Descent~(SGD) optimizer with a weight decay of 0.0001 and a momentum of 0.9.
Unless specified, 
the models are trained for 12 epochs~(1$\times$ learning schedule) and the initial learning rate is set to 0.02 and then reduced by a factor of 10 at the $8$-th epoch and the $11$-th epoch. 
This detector training schedule is used for both the na\"ive alignment (Section~\ref{sec:naive}) and self-training (Section\ref{visual-representation}).
In terms of the data augmentation for the na\"ive alignment, 
we use 640-800 scale jittering and horizontal flipping.\\[-8pt]

{\noindent \bf Regional prompt learning. }
We train the learnable prompt vectors for 6 epochs.
Empirically, we find that the model is not sensitive to the number of prompt vectors, we therefore use two vectors, 
one before $g(\underline{\text{category}})$ as a prefix vector,
and one after $g(\underline{\text{category}})$ as a suffix vector. \\[-8pt]

{\noindent \bf One-iteration prompt learning and image sourcing.}
For the first iteration, we train the prompt using the image crops from LVIS-base. 
Specifically, we expand the ground truth box to triple the height or width for each side, and take crops from the images based on the extended bounding box. 
Then we randomly select up to 200 image crops for each base class to train the prompt vectors. 
At the stage of image sourcing, we search for the web images on LAION-400M using the learned prompt via the \emph{knn} indices of LAION-400M, 
and download about 300 images for each novel class, 
forming LAION-novel for later self-training. \\[-8pt]

{\noindent \bf Multi-iteration prompt learning and image sourcing.}
If we perform the prompt learning and image sourcing for more than one iteration, we start the prompt learning using LVIS-base, 
and search more images for base categories from LAION-400M.
Combining the original LVIS-base and newly sourced images,
we can again update the prompt vectors, 
and used to search images for novel categories later on,
constructing the external LAION-novel dataset.\\[-8pt]

{\noindent \bf Self-training.}
We first train the detector using the LVIS-base images for 6 epochs, 
and then train on both LVIS-base and LAION-novel for another 6 epochs. 
For the LAION-novel images, 
they are often object-centric due to the regional prompt learning,
we use a smaller resolution and do 160$\sim$800 scale jittering.
To guarantee high-quality of pseudo labels, 
we adopt a multi-scale inference scheme to generate the pseudo ground truth bounding boxes for images from LAION-novel.
Specifically, one image scale is randomly selected from 160$\sim$360, 
and the other is randomly selected from 360$\sim$800. 
The two images of different scales are fed into the detector, 
and one pseudo bounding box is generated for each of them.
We select the most confident prediction from both images as the final pseudo bounding boxes, and use them to further self-train the detector.\\[-8pt]

{\noindent \bf Training for more epochs.}
To compare with state-of-the-art detectors,
we train the models with a batchsize of 64 on 8 GPUs, 
for 72 epochs~(6$\times$ learning schedule), with 100$\sim$1280 scale jittering.

\subsection{Ablation Study}
\label{sec:ablation}
In this section, 
we conduct ablation studies on the LVIS dataset,
to thoroughly validate the effectiveness of the proposed components,
including the RPL, iterative candidate sourcing, and self-training. 
In addition, as for studying other hyper-parameters, 
we also conduct comparison experiment to other heuristics for box selection,
effect of sourced candidate images, and finally on the training detail for 
whether to update the class-agnostic region proposal during self-training.\\[-8pt]

{\noindent \bf Regional prompt learning~(RPL). } 
To demonstrate the effectiveness of training open-vocabulary detector, 
by aligning the visual and textual latent space, 
we compare the learned prompt with the manual prompt.
For simplicity, we only use two learnable vectors~(one for prefix, 
and one for suffix) in RPL. 
When using more prompt vectors, we did not observe clear benefits.
 
As shown in Table~\ref{tab:prmpt-learning}, 
we first investigate the performance with the manual prompt of 
``{a photo of [\underline{\text{category}}]}'',
which has also been used in previous works~\cite{zhou2021learning,gu2021open,zhou2022detecting}. 
However, it only brings a limited generalisation, 
yielding a $7.4$ AP on novel categories;
Secondly, after adding more detailed description to the prompt template, 
{\em i.e.}~use the ``a photo of [\underline{\text{category}}], which is [\underline{\text{description}}]'',
the lexical ambiguity can be alleviated, 
and lead to an improvement of $1.6$ AP on novel categories;
Lastly, we verify the effectiveness of our proposed prompt learning,
which further brings a performance improvement by $3.7$ AP and $2.1$ AP on novel categories, comparing to the two manual prompts respectively.\\[-8pt]

\begin{table}[!htb]
\vspace{-0.6cm}
\caption{Comparison on manually designed and learned prompt.
Here, we only use two learnable prompt vectors in PRL, 
{\em i.e.}~$[1+1]$ refers to using one vector for prefix, and one vector for suffix.}
\small
\setlength\tabcolsep{4pt}
\begin{center}
\begin{tabular}{lccccc}
\hline\noalign{\smallskip}
& Prompt & AP$_{novel}$ & AP$_{c}$ & AP$_{f}$ & AP \\
\noalign{\smallskip}
\hline
\noalign{\smallskip}
``a photo of [\underline{\text{category}}]'' & manual & 7.4 & 17.2 & 26.1 & 19.0 \\
``a photo of [\underline{\text{category}}], 
which is [\underline{\text{description}}]'' & manual & 9.0 & 18.6 & 26.5 & 20.1 \\
regional prompt learning & [1+1] & 11.1 & 18.8 & 26.6 & 20.3 \\ \bottomrule
\end{tabular}
\label{tab:prmpt-learning}
\vspace{-0.6cm}
\end{center}
\end{table}

{\noindent \bf Self-training. } 
We evaluate the performance of after self-training the detector,
both with and without the iterative candidate image sourcing.
As shown in Table~\ref{tab:extract-data}, 
it can always bring a noticeable improvement with different prompts, 
for example, from $9.0$ to $15.3$ AP for manual prompt, 
and $11.1$ to $15.9$ AP for learnt prompt.
Additionally, 
while conducting a second-round regional prompt learning and image sourcing,
our proposed PromptDet really shines, 
significantly outperforming the manually designed prompt,
reaching $19.0$ AP on novel categories.
It demonstrates the effectiveness of self-training and iterative prompt learning for sourcing higher quality images. 
We also conduct a third-round regional prompt learning, and it yields $19.3$ AP on novel categories. For simplicity, we iterate the prompt learning twice in the following experiments.\\[-8pt]

\begin{table}[!htb]
\caption{Effectiveness of self-training with different prompts.
1-iter, 2-iter and 3-iter denote that Stage-I (\ie RPL) and Stage-II (\ie image sourcing) are performed for one, two or three iterations, respectively.
}
\vspace{-0.2cm}
\small
\setlength\tabcolsep{2.8pt}
\begin{center}
\begin{tabular}{lccccc}
\hline\noalign{\smallskip}
Prompt method &Self-training & AP$_{novel}$ & AP$_{c}$ & AP$_{f}$ & AP \\
\noalign{\smallskip}
\hline
\noalign{\smallskip}
\multirow{2}*{``a photo of [\underline{\text{category}}], which is [\underline{\text{description}}]''} & & 9.0 & 18.6 & 26.5 & 20.1 \\
& $\checkmark$ & 15.3 & 17.7 & 25.8 & 20.4 \\
\noalign{\smallskip}
\hline
\noalign{\smallskip}
Regional prompt learning & & 11.1 & 18.8 & 26.6 & 20.3 \\
PromptDet~(1-iter) & $\checkmark$ & 15.9 & 17.6 & 25.5 & 20.4 \\ 
PromptDet~(2-iter) & $\checkmark$ & 19.0 & 18.5 & 25.8 & 21.4 \\
PromptDet~(3-iter) & $\checkmark$ & 19.3 & 18.3 & 25.8 & 21.4 
\\ \bottomrule
\end{tabular}
\label{tab:extract-data}
\vspace{-0.2cm}
\end{center}
\end{table}

\begin{figure*}[!htb]
	\centering
	\includegraphics[width=\textwidth]{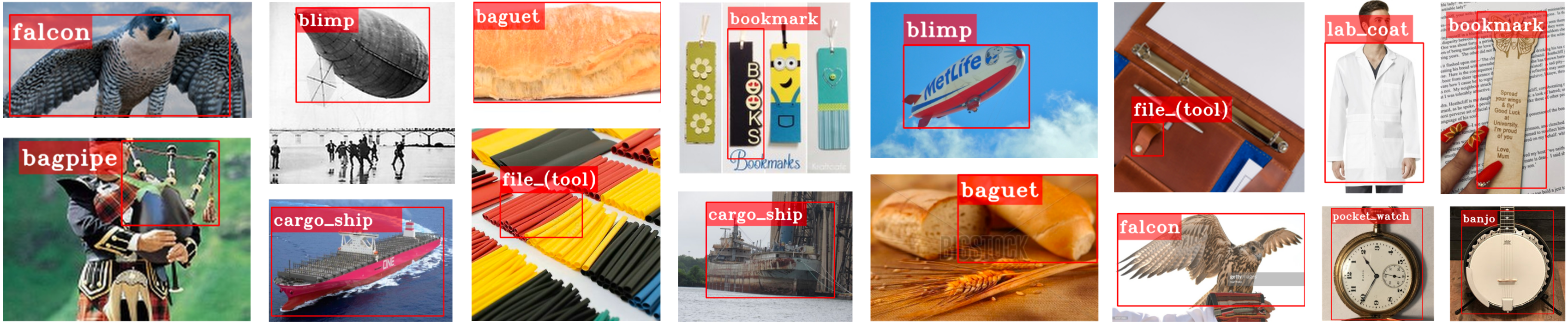}
	\vspace{-0.4cm}
	\caption{Visualisation of the generated pseudo ground truth for the sourced images.}
	\vspace{-0.3cm}
	\label{pseudo-labeling-images}
\end{figure*}

{\noindent \bf Box generation. } 
As for pseudo labeling the boxes, 
we show some visualisation examples by taking the most confident predictions on the sourced candidate images, as shown in Figure~\ref{pseudo-labeling-images}.

Quantitatively, 
we compare with three different heuristic strategies for box generation,
as validated in Detic~\cite{zhou2022detecting}: 
(1) use the whole image as proposed box; 
(2) the proposal with max size; 
(3) the proposal with max RPN score. 

As shown in Table~\ref{tab:ablation}~(left), 
we observe that using the most confident predictions as pseudo ground truth significantly outperforms the other strategies.
Specifically, we conjecture this performance gap between ours and the max-size boxes~(used in Detic) might be due to the difference on the external data.
Detic exploits ImageNet21K with images being manually verified by human annotators, however, we only adopt the noisy, uncurated web images,
training on the bounding boxes generated by heuristic may thus incur erroneous supervision in the detector training. \\[-8pt]

{\noindent \bf Sourcing variable candidate images. }
We investigate the performance variation while increasing the number of uncurated web images for self-training.
As shown in Table~\ref{tab:ablation}~(right), 
$0$ image denotes the training scenario with no self-training involved, 
and when increasing the number of sourced images from 50 to 300, 
the performance tends to be increasing monotonically,
from $14.6$ to $19.0$ AP on novel categories,
showing the scalability of our proposed self-training mechanism.
However, we found the LAION-400M dataset can only support at most 300 images for most categories, 
and sourcing more images would require to use a larger corpus,
we leave this as a future work.\\[-8pt]

\begin{table*}[!htb]
\caption{\textbf{Left}: the comparison on different box generation methods.
\textbf{Right}: the effect on increasing the sourced candidate images.}
\small
\setlength\tabcolsep{1.3pt}
\begin{minipage}[t]{.5\linewidth}
\centering
\begin{tabular}{lcccc}
\hline\noalign{\smallskip}
Method & AP$_{novel}$ & AP$_{c}$ & AP$_{f}$ & AP \\
\noalign{\smallskip}
\hline
\noalign{\smallskip}
w/o self-training & 10.4 & 19.5 & 26.6 & 20.6 \\
\noalign{\smallskip}
\hline
\noalign{\smallskip}
image & 9.9 & 18.8 & 26.0 & 20.1 \\
max-size & 9.5 & 18.8 & 26.1 & 20.1 \\
max-obj.-score & 11.3 & 18.7 & 26.0 & 20.3 \\
max-pred.-score~(ours) & 19.0 & 18.5 & 25.8 & 21.4 \\ \bottomrule
\end{tabular}
\end{minipage} 
\hfill
\hspace{12pt}
\begin{minipage}[t]{.47\linewidth}
\centering
\setlength\tabcolsep{1.3pt}
\begin{tabular}{ccccc}
\hline\noalign{\smallskip}
\#Web images & AP$_{novel}$ & AP$_{c}$ & AP$_{f}$ & AP \\
\noalign{\smallskip}
\hline
\noalign{\smallskip}
0 & 10.4 & 19.5 & 26.6 & 20.6 \\
50 &  14.6 & 19.3 & 26.2 & 21.2 \\
100 & 15.8 & 19.3 & 26.2 & 21.4 \\
200 & 17.4 & 19.1 & 26.0 & 21.5 \\
300 & 19.0 & 18.5 & 25.8 & 21.4 \\ \bottomrule
\end{tabular}
\end{minipage}
\vspace{-0.2cm}
\label{tab:ablation}
\end{table*}

{\noindent \bf Updating class-agnostic RPN and box head. } 
Here, we conduct the ablation study on updating or freezing the class-agnostic RPN or box regression during self-training.
As shown in Table~\ref{tab:ablation2}~(left), 
we find that freezing these two components can be detrimental, 
leading to a 1.8 AP~(from 19.0 to 17.2) performance drop on detecting the novel categories.\\[-8pt]

{\noindent \bf Top-K proposals for pseudo labeling. } 
We investigate the performance by varying the number of box proposal in the pseudo-labeling procedure. 
As shown in Table~\ref{tab:ablation2}~(right), 
selecting the most confident prediction among the top-20 proposals yields the best performance, and taking all object proposals presents the worst performance~(10.4 AP $vs.$ 19.0 AP) for novel categories.
The other options are all viable, though with some performance drop.
We set $\text{K} = 20$ for our experiments.

\begin{table*}[!htb]
\vspace{-0.5cm}
\caption{
\textbf{Left}: the  
ablation study on updating class-agnostic RPN and box regression during self-training.
\textbf{Right}: the analysis on the effect of generating variable pseudo boxes from RPN.}
\small
\begin{minipage}[t]{.47\linewidth}
\setlength\tabcolsep{1.5pt}
\centering
\begin{tabular}{cccccc}
\hline\noalign{\smallskip}
RPN classifier & Box head & AP$_{novel}$ & AP$_{c}$ & AP$_{f}$ & AP \\
\noalign{\smallskip}
\hline
\noalign{\smallskip}
 &  & 17.2 & 18.2 & 25.8 & 21.0  \\
$\checkmark$ & & 18.1 & 18.3 & 25.7 & 21.2 \\
$\checkmark$ & $\checkmark$ & 19.0 & 18.5 & 25.8 & 21.4 \\ \bottomrule
\end{tabular}
\end{minipage} 
\hfill
\hspace{5pt}
\begin{minipage}[t]{.6\linewidth}
\setlength\tabcolsep{1.5pt}
\centering
\begin{tabular}{ccccc}
\hline\noalign{\smallskip}
\#Proposals & AP$_{novel}$ & AP$_{c}$ & AP$_{f}$ & AP \\
\noalign{\smallskip}
\hline
\noalign{\smallskip}
10 &  17.2 & 18.7 & 25.7 & 21.2 \\
20 &  19.0 & 18.5 & 25.8 & 21.4 \\
30 &  16.1 & 18.8 & 25.9 & 21.1 \\
1000 (all) & 10.4 & 19.5 & 26.6 & 20.6 \\ \bottomrule
\end{tabular}
\end{minipage}
\label{tab:ablation2}
\vspace{-0.4cm}
\end{table*}

\subsection{Comparison with the State-of-the-Art}
In Table~\ref{results},
we compare the proposed method with other open-vocabulary object detectors~\cite{gu2021open,zhou2022detecting} on the LIVS v1.0 validation set. Limited by the computational resource, 
our best model is only trained for 72 epochs, 
and achieving $21.4$ AP for the novel categories, 
surpassing the recent state-of-the-art ViLD-ens~\cite{gu2021open} and Detic~\cite{zhou2022detecting} by $4.8$ AP and $3.6$ AP respectively.
Additionally, we observe that training for longer schedule can significantly improve the detection performance on {\em common} and {\em frequent} categories,
from $18.5$ AP to $23.3$ AP 
and $25.8$ AP to $29.3$ AP respectively.

\begin{table}[!htb]
\caption{Detection results on the LVIS v1.0 validation set. 
Both Detic and our proposed approach have exploited the external images.
However, in Detic, the images are manually annotated and 
thus indicated by `*'.
Notably, PromptDet does not require a knowledge distillation from the CLIP visual encoder at the detector training, which is shown to prominently boost the performance but significantly increase the training costs.
}
\begin{center}
\small
\setlength\tabcolsep{2.8pt}
\begin{tabular}{lcccccccc}
\hline\noalign{\smallskip}
\multirow{2}*{Method} & \multirow{2}*{Epochs} & Scale & Input & \multirow{2}*{\#External} & \multirow{2}*{AP$_{novel}$} & \multirow{2}*{AP$_{c}$} & \multirow{2}*{AP$_{f}$} & \multirow{2}*{AP} \\
&  & Jitter & Size &  &  \\
\noalign{\smallskip}
\hline
\noalign{\smallskip}
ViLD-text~\cite{gu2021open} & 384 & 100$\sim$2048 & 1024$\times$1024 & 0 & 10.1 & 23.9 & 32.5 & 24.9 \\
ViLD~\cite{gu2021open} & 384 & 100$\sim$2048 & 1024$\times$1024 & 0  & 16.1 & 20.0 & 28.3 & 22.5  \\
ViLD-ens.~\cite{gu2021open} & 384& 100$\sim$2048 & 1024$\times$1024 & 0& 16.6 & 24.6 & 30.3 & 25.5  \\
Detic~\cite{zhou2022detecting} & 384 & 100$\sim$2048 & 1024$\times$1024 & 1.2M* & 17.8 & 26.3 & 31.6 & 26.8 \\ \midrule
PromptDet & 12 & 640$\sim$800 & 800$\times$800 & 0.1M & 19.0 & 18.5 & 25.8 & 21.4 \\ 
PromptDet & 72 & 100$\sim$1280 & 800$\times$800 & 0.1M & {\bf 21.4} & 23.3 & 29.3 & 25.3 \\ \bottomrule
\end{tabular}
\label{results}
\vspace{-0.8cm}
\end{center}
\end{table}

Additionally, we compare with previous works on open-vocabulary COCO benchmark. Following~\cite{bansal2018zero,zhou2022detecting}, we apply the 48/17 base/novel split setting on MS-COCO, and report the box Average Precision at the IoU threshold 0.5. As Table~\ref{open-coco} shows, PromptDet trained for 24 epochs outperforms Detic on both novel-class mAP (26.6 AP \emph{vs.}~24.1 AP) and overall mAP (50.6 AP \emph{vs.}~44.7 AP) with the same input image resolution (\ie~640$\times$640).

\vspace{-0.7cm}
\begin{table}[!htb]
\setlength\tabcolsep{3.8pt}
\caption{Results on open-vocabulary COCO. 
Numbers are copied from~\cite{zhou2022detecting}}
\centering
\begin{tabular}{lcccc}
\hline\noalign{\smallskip}
Method & Epochs & Input size & AP50$_{\text{novel}}^{\text{box}}$ & AP50$_{\text{all}}^{\text{box}}$ \\
\noalign{\smallskip}
\hline
\noalign{\smallskip}
WSDDN~\cite{bilen2016weakly} & 96 & 640$\times$640 & 5.9 & 39.9 \\
DLWL~\cite{ramanathan2020dlwl} & 96 & 640$\times$640 & 19.6 & 42.9 \\
Predicted~\cite{redmon2017yolo9000} & 96 & 640$\times$640 & 18.7 & 41.9 \\
Detic~\cite{zhou2022detecting} & 96 & 640$\times$640 & 24.1 & 44.7 \\
\noalign{\smallskip}
\hline
\noalign{\smallskip}
PromptDet & 24 & 640$\times$640 & \textbf{26.6} & 50.6 \\ 
\bottomrule
\end{tabular}
\label{open-coco}
\vspace{-0.2cm}
\end{table}

\section{Conclusion}
In this work, we propose an open-vocabulary object detector PromptDet, 
which is able to detect novel categories without any manual annotations. Specifically, we first use the pretrained, frozen CLIP text encoder, as an ``off-the-shelf'' classifier generator in two-stage object detector. 
Then we propose a regional prompt learning method to steer the textual latent space towards the task of object detection,
{\em i.e.,}~transform the textual embedding space, to better align the visual representation of object-centric images.
In addition, we further develop a self-training regime, 
which enables to iteratively high-quality source candidate images from a large corpus of uncurated, external images, and self-train the detector.
With these improvements, PromptDet achieved a 21.4 AP of novel classes on LVIS, surpassing the state-of-the-art open-vocabulary object detectors by a large margin, with much lower training costs.

\clearpage
%
%
\bibliographystyle{splncs04}
\bibliography{egbib}

\begin{thebibliography}{10}
\providecommand{\url}[1]{\texttt{#1}}
\providecommand{\urlprefix}{URL }
\providecommand{\doi}[1]{https://doi.org/#1}

\bibitem{akata2016multi}
Akata, Z., Malinowski, M., Fritz, M., Schiele, B.: Multi-cue zero-shot learning
  with strong supervision. In: Proceedings of the IEEE Conference on Computer
  Vision and Pattern Recognition. pp. 59--68 (2016)

\bibitem{bansal2018zero}
Bansal, A., Sikka, K., Sharma, G., Chellappa, R., Divakaran, A.: Zero-shot
  object detection. In: Proceedings of the European Conference on Computer
  Vision. pp. 384--400 (2018)

\bibitem{bilen2016weakly}
Bilen, H., Vedaldi, A.: Weakly supervised deep detection networks. In: cvpr.
  pp. 2846--2854 (2016)

\bibitem{cacheux2019modeling}
Cacheux, Y.L., Borgne, H.L., Crucianu, M.: Modeling inter and intra-class
  relations in the triplet loss for zero-shot learning. In: Proceedings of the
  IEEE Conference on Computer Vision and Pattern Recognition. pp. 10333--10342
  (2019)

\bibitem{elhoseiny2017link}
Elhoseiny, M., Zhu, Y., Zhang, H., Elgammal, A.: Link the head to the" beak":
  Zero shot learning from noisy text description at part precision. In:
  Proceedings of the IEEE Conference on Computer Vision and Pattern
  Recognition. pp. 5640--5649 (2017)

\bibitem{everingham2015pascal}
Everingham, M., Eslami, S., Van~Gool, L., Williams, C.K., Winn, J., Zisserman,
  A.: The pascal visual object classes challenge: A retrospective.
  International Journal of Computer Vision  \textbf{111}(1),  98--136 (2015)

\bibitem{fan2020few}
Fan, Q., Zhuo, W., Tang, C.K., Tai, Y.W.: Few-shot object detection with
  attention-rpn and multi-relation detector. In: Proceedings of the IEEE
  Conference on Computer Vision and Pattern Recognition. pp. 4013--4022 (2020)

\bibitem{feng2021tood}
Feng, C., Zhong, Y., Gao, Y., Scott, M.R., Huang, W.: Tood: Task-aligned
  one-stage object detection. In: Proceedings of the International Conference
  on Computer Vision. pp. 3490--3499. IEEE Computer Society (2021)

\bibitem{feng2021exploring}
Feng, C., Zhong, Y., Huang, W.: Exploring classification equilibrium in
  long-tailed object detection. In: Proceedings of the International Conference
  on Computer Vision. pp. 3417--3426 (2021)

\bibitem{frome2013devise}
Frome, A., Corrado, G.S., Shlens, J., Bengio, S., Dean, J., Ranzato, M.,
  Mikolov, T.: Devise: A deep visual-semantic embedding model. Advances in
  neural information processing systems  \textbf{26} (2013)

\bibitem{gu2021open}
Gu, X., Lin, T.Y., Kuo, W., Cui, Y.: Open-vocabulary object detection via
  vision and language knowledge distillation. arXiv preprint arXiv:2104.13921
  (2021)

\bibitem{gupta2019lvis}
Gupta, A., Dollar, P., Girshick, R.: Lvis: A dataset for large vocabulary
  instance segmentation. In: Proceedings of the IEEE Conference on Computer
  Vision and Pattern Recognition. pp. 5356--5364 (2019)

\bibitem{he2017mask}
He, K., Gkioxari, G., Doll{\'a}r, P., Girshick, R.: Mask r-cnn. In: Proceedings
  of the International Conference on Computer Vision. pp. 2961--2969 (2017)

\bibitem{ji2018stacked}
Ji, Z., Fu, Y., Guo, J., Pang, Y., Zhang, Z.M., et~al.: Stacked
  semantics-guided attention model for fine-grained zero-shot learning.
  Advances in Neural Information Processing Systems  \textbf{31} (2018)

\bibitem{jia2021scaling}
Jia, C., Yang, Y., Xia, Y., Chen, Y.T., Parekh, Z., Pham, H., Le, Q., Sung,
  Y.H., Li, Z., Duerig, T.: Scaling up visual and vision-language
  representation learning with noisy text supervision. In: Proceedings of the
  International Conference on Machine Learning. pp. 4904--4916. PMLR (2021)

\bibitem{kang2019few}
Kang, B., Liu, Z., Wang, X., Yu, F., Feng, J., Darrell, T.: Few-shot object
  detection via feature reweighting. In: Proceedings of the International
  Conference on Computer Vision. pp. 8420--8429 (2019)

\bibitem{Kaul22}
Kaul, P., Xie, W., Zisserman, A.: Label, verify, correct: A simple few shot
  object detection method. In: Proceedings of the IEEE Conference on Computer
  Vision and Pattern Recognition (2022)

\bibitem{Li19}
Li, Z., Yao, L., Zhang, X., Wang, X., Kanhere, S., Zhang, H.: Zero-shot object
  detection with textual descriptions. In: Proceedings of the AAAI Conference
  on Artificial Intelligence (2019)

\bibitem{lin2017focal}
Lin, T.Y., Goyal, P., Girshick, R., He, K., Doll{\'a}r, P.: Focal loss for
  dense object detection. In: Proceedings of the International Conference on
  Computer Vision. pp. 2980--2988 (2017)

\bibitem{lin2014microsoft}
Lin, T.Y., Maire, M., Belongie, S., Hays, J., Perona, P., Ramanan, D.,
  Doll{\'a}r, P., Zitnick, C.L.: Microsoft coco: Common objects in context. In:
  Proceedings of the European Conference on Computer Vision. pp. 740--755
  (2014)

\bibitem{Mori99}
Mori, Y., Takahashi, H., Oka, R.: Image-to-word transformation based on
  dividing and vector quantizing images with words. In: MISRM (1999)

\bibitem{radford2021learning}
Radford, A., Kim, J.W., Hallacy, C., Ramesh, A., Goh, G., Agarwal, S., Sastry,
  G., Askell, A., Mishkin, P., Clark, J., et~al.: Learning transferable visual
  models from natural language supervision. In: Proceedings of the
  International Conference on Machine Learning. pp. 8748--8763. PMLR (2021)

\bibitem{Rahman20}
Rahman, S., Khan, S., Barnes, N.: Improved visual-semantic alignment for
  zero-shot object detection. In: Proceedings of the AAAI Conference on
  Artificial Intelligence (2020)

\bibitem{ramanathan2020dlwl}
Ramanathan, V., Wang, R., Mahajan, D.: Dlwl: Improving detection for lowshot
  classes with weakly labelled data. In: cvpr. pp. 9342--9352 (2020)

\bibitem{redmon2017yolo9000}
Redmon, J., Farhadi, A.: Yolo9000: better, faster, stronger. In: cvpr. pp.
  7263--7271 (2017)

\bibitem{ren2015faster}
Ren, S., He, K., Girshick, R., Sun, J.: Faster r-cnn: Towards real-time object
  detection with region proposal networks. arXiv preprint arXiv:1506.01497
  (2015)

\bibitem{rohrbach2011evaluating}
Rohrbach, M., Stark, M., Schiele, B.: Evaluating knowledge transfer and
  zero-shot learning in a large-scale setting. In: Proceedings of the IEEE
  Conference on Computer Vision and Pattern Recognition (2011)

\bibitem{schuhmann2021laion}
Schuhmann, C., Vencu, R., Beaumont, R., Kaczmarczyk, R., Mullis, C., Katta, A.,
  Coombes, T., Jitsev, J., Komatsuzaki, A.: Laion-400m: Open dataset of
  clip-filtered 400 million image-text pairs. arXiv preprint arXiv:2111.02114
  (2021)

\bibitem{tian2019fcos}
Tian, Z., Shen, C., Chen, H., He, T.: Fcos: Fully convolutional one-stage
  object detection. In: Proceedings of the International Conference on Computer
  Vision. pp. 9627--9636 (2019)

\bibitem{Weston11}
Weston, J., Bengio, S., Usunier, N.: Wsabie: Scaling up to large vocabulary
  image annotation. In: IJCAI (2011)

\bibitem{xie2021zsd}
Xie, J., Zheng, S.: Zsd-yolo: Zero-shot yolo detection using vision-language
  knowledgedistillation. arXiv preprint arXiv:2109.12066  (2021)

\bibitem{Zareian21}
Zareian, A., Rosa, K.D., Hu, D.H., Chang, S.F.: Open-vocabulary object
  detection using captions. In: Proceedings of the IEEE Conference on Computer
  Vision and Pattern Recognition (2021)

\bibitem{zhao2017open}
Zhao, H., Puig, X., Zhou, B., Fidler, S., Torralba, A.: Open vocabulary scene
  parsing. In: Proceedings of the International Conference on Computer Vision.
  pp. 2002--2010 (2017)

\bibitem{zhong2020representation}
Zhong, Y., Deng, Z., Guo, S., Scott, M.R., Huang, W.: Representation sharing
  for fast object detector search and beyond. In: Proceedings of the European
  Conference on Computer Vision. pp. 471--487. Springer (2020)

\bibitem{zhou2021learning}
Zhou, K., Yang, J., Loy, C.C., Liu, Z.: Learning to prompt for vision-language
  models. arXiv preprint arXiv:2109.01134  (2021)

\bibitem{zhou2022detecting}
Zhou, X., Girdhar, R., Joulin, A., Kr{\"a}henb{\"u}hl, P., Misra, I.: Detecting
  twenty-thousand classes using image-level supervision. arXiv preprint
  arXiv:2201.02605  (2022)

\end{thebibliography}

\clearpage


\section*{Supplementary material}

# – Supplementary material –
# PromptDet: Towards Open-vocabulary Detection using Uncurated Images

Chengjian Feng[1], Yujie Zhong[1], Zequn Jie[1], Xiangxiang Chu[1]
Haibing Ren[1], Xiaolin Wei[1], Weidi Xie[2, ✉], and Lin Ma[1]

[1]Meituan Inc.   [2]Shanghai Jiao Tong University

## 1  Open-vocabulary COCO benchmark details

Standard MS-COCO [3] contains 80 categories. Following [1, 4], we only select 48 classes as base classes, and 17 classes as novel classes. The *train* set and *minival* set are the same as the standard MS-COCO 2017, however, only images containing at least one base class are used for the training. For self-training, we search for the images from LAION-400M and download about 1500 images for each novel category. We summarise the dataset statistics for open-vocabulary COCO benchmark in Table 1.

Since there are no class descriptions in the meta data of MS-COCO, we use the descriptions from LVIS if the category is convered by LVIS. For the other categories, we get their descriptions by searching on Google, for instance, {category: *"donut"*, description: *"a small fried cake of sweetened dough, typically in the shape of a ball or ring"*}.

We conduct the experiment using Mask-RCNN [2] with a ResNet-50-FPN backbone. The model is trained with a batchsize of 64 on 8 GPUs, for 24 epochs (2x learning schedule). Similar to open-vocabulary LVIS benchmark, we use COCO-base to train an initial open-vocabulary object detector. Then we use a combination of COCO-base and LAION-novel datasets for self-training. Considering the small number of the base categories on open-vocabulary COCO benchmark, we select 1000 images for each base category during the regional prompt learning, and set K = 10 to retain the precise bounding boxes for the pseudo labeling. Besides, the COCO-base images contain abundant objects of the novel categories. To enhance the recall of the region proposal network (RPN) for the novel categories, we compute the similarity between the features from the negative box proposals and the text embedding of novel categories, and treat the box proposals with high similarity (*i.e.* lager than 0.1) as the positive samples, during the training of RPN.

## 2  Cross-dataset transfer

To demonstrate the generalisation, we directly evaluate the detector from open-vocabulary LVIS benchmark on MS-COCO [3], *without* any finetuning on it.

Table 1: A summary of dataset statistics for open-vocabulary COCO benchmark. The numbers in bracket refer to the number of base and novel categories.

| Dataset | Train | Eval. | Definition | #Images | #Categories |
|---|---|---|---|---|---|
| MS-COCO | – | – | original MS-COCO dataset | 0.1M | 80 |
| LAION-400M | – | – | image-text pairs filtered by CLIP | 400M | unlabeled |
| COCO-base | ✓ | ✗ | base categories on MS-COCO | 0.1M | 48 |
| LAION-novel | ✓ | ✗ | image subset of novel categories | 25K | 17 (noisy) |
| COCO *minival* | ✗ | ✓ | MS-COCO validation set | 5K | 65 (48+17) |

Specifically, MS-COCO contains 80 categories, with only 59 categories covered by LVIS-base (denoted as base categories, $\mathcal{C}_{base}$). Therefore, the other 21 categories can be used to benchmark the generalisation ability of PromptDet.

As shown in Table 2, the learned regional prompt improves the generalisation of the detector, and achieves 29.0 AP for the novel categories, suppressing the manual prompt by 1.6 AP (27.4 AP *vs*. 29.0 AP). Similarly, with more training (72 epochs), the model reaches 31.4 AP on detecting the 21 novel categories on MS-COCO.

Table 2: Generalisation from LVIS-base to MS-COCO with different prompts. The trained detector is directly transferred to MS-COCO by replacing the category and description embedding in the classifiers without fine-tuning.

|  | Prompt | Epochs | $AP_{novel}$ | $AP_{base}$ | AP |
|---|---|---|---|---|---|
| "a photo of [category], which is [description]" | manual | 12 | 27.4 | 26.9 | 27.0 |
| PromptDet | learnable | 12 | 29.0 | 26.5 | 27.2 |
| PromptDet | learnable | 72 | 31.4 | 29.1 | 29.7 |

## 3   Qualitative Results

In Figure 1, we show the qualitative results from PromptDet on open-vocabulary LVIS benchmark. Without any manual annotations, the detector can now accurately localise and recognise the objects from a diverse set of novel categories.

Fig. 1: Qualitative results from our PromptDet on images from LVIS validation set. The boxes with green denote the objects from novel categories, while blue boxes refer to the objects from base categories.

# References

1. Bansal, A., Sikka, K., Sharma, G., Chellappa, R., Divakaran, A.: Zero-shot object detection. In: Proceedings of the European Conference on Computer Vision. pp. 384–400 (2018)
2. He, K., Gkioxari, G., Dollár, P., Girshick, R.: Mask r-cnn. In: Proceedings of the International Conference on Computer Vision. pp. 2961–2969 (2017)
3. Lin, T.Y., Maire, M., Belongie, S., Hays, J., Perona, P., Ramanan, D., Dollár, P., Zitnick, C.L.: Microsoft coco: Common objects in context. In: Proceedings of the European Conference on Computer Vision. pp. 740–755 (2014)
4. Zhou, X., Girdhar, R., Joulin, A., Krähenbühl, P., Misra, I.: Detecting twenty-thousand classes using image-level supervision. arXiv preprint arXiv:2201.02605 (2022)



\end{document}